\documentclass[a4paper, 12pt]{article}

\usepackage{amsmath, amssymb, multirow}
\usepackage[dvips]{graphicx}

\title{Network of two-Chinese-character compound words
in Japanese language}

\author{Ken Yamamoto and Yoshihiro Yamazaki}

\newcommand{\institution}[1]{\date{\it\small #1}}
\institution{Department of Physics, Waseda University, Tokyo, 169-8555, Japan}

\newcommand{\ang}[1]{\langle #1 \rangle}

\begin{document}
\maketitle

% Abstract
\begin{abstract}
Some statistical properties of a network
of two-Chinese-character compound words
in Japanese language are reported.
In this network,
a node represents a Chinese character and
an edge represents a two-Chinese-character compound word.
It is found that this network has properties of ``small-world''
and ``scale-free.''
A network formed by only Chinese characters for common use
({\it joyo-kanji} in Japanese),
which is regarded as a subclass of the
original network, also has
small-world property.
However, a degree distribution of the network
exhibits no clear power law.
In order to reproduce disappearance of the power-law property,
a model for a selecting process of
the Chinese characters for common use
is proposed.
\end{abstract}

\section{Introduction}
% Short review about network science
It has been found that
a great variety of systems,
such as internet \cite{Faloutsos, Rosato},
collaboration in science \cite{Newman1, Newman2},
food web \cite{Dunne, Gilbert},
have network structures;
systems consist of a group of nodes which
interact mutually through edges.
Network science supplies some methods
to understand topological structures of such systems.
Recently, it has been proved that the properties of
small-world \cite{Watts, Strogatz} and scale-free \cite{Barabasi}
are important and 
that
many networks share these properties.
For typical examples,
human languages
have been modeled 
in the framework of complex networks
so as to investigate graphemic \cite{Li},
phonetic \cite{Soares}, syntactic \cite{Zhou} and 
semantic \cite{Steyvers} structures.

% Chinese character in Japanese language
Chinese characters are main elements
in the writing system of Japanese language.
One of the most remarkable features of
Chinese characters is that they are
ideograms, that is,
a single Chinese character can convey
its own meaning.

% two-character compound words in Japanese language
Japanese language possesses many words
constructed by combining two Chinese characters.
Such words are called `two-Chinese-character compound words'
({\it niji-jukugo} in Japanese),
and we adopt the name `two-character compounds' hereafter.
For instance,
in the Japanese-language dictionary {\it Kojien} \cite{dictionaries},
about 90,000 words among about 200,000 headwords
are two-character compounds.
So far, researches on two-character compounds
in Japanese language
have been concentrated mostly on morphological structures
\cite{Joyce, HMasuda}
and cognitive processes \cite{Morita, Tamaoka}.
However, studies of
the two-character compounds in Japanese language
based on the network science seem to be insufficient.
In the present paper,
we report analysis results of networks
of two-character compounds
in Japanese language.

\section{Method}
First, we extracted networks
of two-character compounds from the 
following Japanese-language dictionaries:
{\it Kojien}, {\it Iwanami Kokugo Jiten},
{\it Sanseido Kokugo Jiten},
and {\it Mitsumura Kokugo Gakushu Jiten} \cite{dictionaries}.
It is noted that
{\it Kojien}, {\it Iwanami},
and {\it Sanseido} are standard dictionaries,
but {\it Mitsumura} is a dictionary for students
of elementary and junior high school.
We picked out two-character compounds
from the headwords of each dictionary.

\begin{figure}[htb]
\centering
\includegraphics[width=0.8\textwidth]{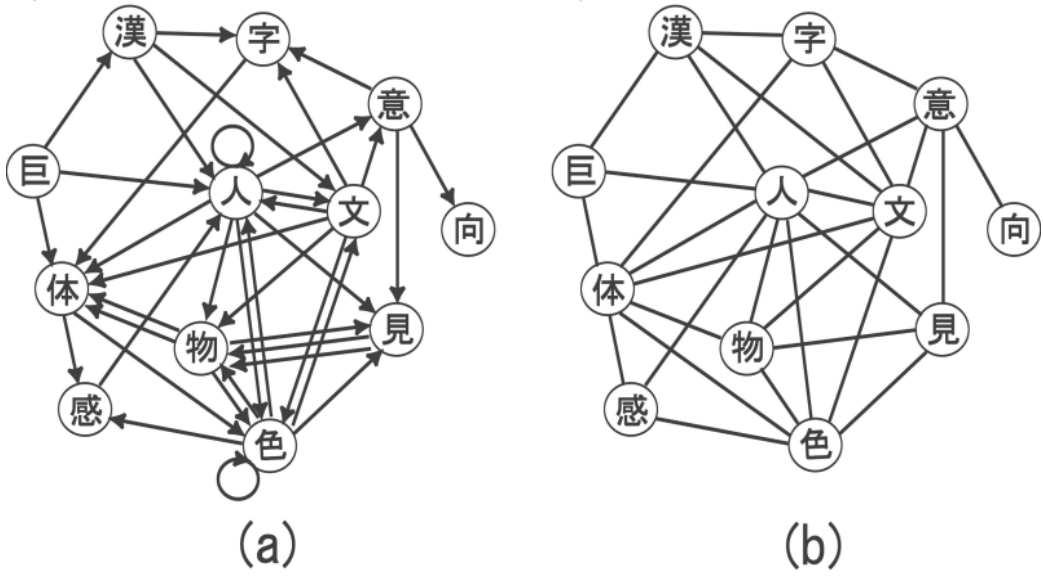}
\caption{K. Yamamoto and Y. Yamazaki}
\label{fig:part}
\end{figure}

In the network of two-character compounds,
each Chinese character corresponds to a node,
and each two-character compound
formed by connecting two nodes is regarded as an edge.
Each edge have a direction
from an upper character to a lower character.
Thus, this network is naturally viewed as a directed network
with multiple edges and self loops.
The direction of edges in the network
deeply relates to lexical structure and meaning of
two-character compounds.
The multiplicity of edges represents the following two aspects:
(i) some two-character compounds have two or more readings, and
(ii) some compounds become other existing compounds
when the upper and lower characters are inverted.
A part of this network is depicted in Fig. \ref{fig:part}.

In the networks we obtained,
all nodes are not connective, and
whole network is made up of $169$ ({\it Kojien}), $152$ ({\it Iwanami}),
$142$ ({\it Sanseido}), and $8$ ({\it Mitsumura}) clusters.
In the following analysis,
we consider the maximal cluster
in the network of each dictionary
(more than 90\% of nodes belong to the maximal cluster).
%And we omit the direction of edge, multiple edges, and self loops,
%for concentrating on a simple structure of the networks.
Since essential features of the networks can be described 
even without
the edge direction and multiplicity and self loops,
we focus on the undirected and unweighted networks.

\section{Results}
\begin{figure}[htb]
\centering
\includegraphics[width=\textwidth]{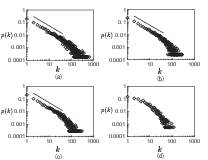}
\caption{K. Yamamoto and Y. Yamazaki}
\label{fig:degree}
\end{figure}

Fundamental results obtained from each dictionary
are summarized in Table \ref{tbl:maximal}.
For instance,
in the case of {\it Kojien},
a pair of two nodes 
is about three steps distant on average,
and at most ten steps distant
(see $\ell$ and $D$ in this Table).
Clustering coefficient $C$ of each network is
about 20 times greater than that of a random network
of the same size in nodes and edges $C_{rand}$.
Therefore, networks of two-character compounds
have short path length and high clustring,
as in many real networks \cite{Albert}.
It is found that
the degree distributions of the three networks
(shown in Fig. \ref{fig:degree} (a)-(c)) display power law
\[
p(k)\propto k^{-\gamma},
\]
where $p(k)$ denotes a fraction of nodes having degree $k$.
Values of $\gamma$ are nearly 1 for these three dictionaries
as shown in Table \ref{tbl:maximal}.
However, as shown in Fig.\ref{fig:degree} (d),
the degree distribution of {\it Mitsumura}
does not exhibit clear power-law property.

\section{Restricted network formed by Chinese characters for common use}
In this section,
we discuss the reason why
the degree distribution of {\it Mitsumura} does not
exhibit power law
(see Fig. \ref{fig:degree} (d) for reference).
There are $1,945$ Chinese characters designated
for common use,
which are called {\it joyo-kanji} in Japanese,
selected by
the Ministry of Education, Science and Culture of Japan
in $1981$.
We call them `common-use characters' heareafter.
The common-use characters are taught during
elementary and junior high school in Japan,
and most Chinese characters used in Japan
are the common-use characters.
Moreover,  
Chinese characters except the common-use characters
are not permitted to use in legal documents.
We next consider a network constructed only
by the common-use characters.
It is noted that this network forms a subclass 
of the original network.

Fundamental results of the network
restricted to the common-use characters are
summarized in Table \ref{tbl:joyo}.
For the first three dictionaries in Table \ref{tbl:joyo},
mean path lengths are small,
and clustering coefficients are large,
compared to those presented in Table \ref{tbl:maximal}.
On the other hand,
the properties of the network of {\it Mitsumura}
in Table \ref{tbl:joyo} is the
same as those in Table \ref{tbl:maximal}.
This reflects that two-character compounds
listed in {\it Mitsumura} are all
constructed from the common-use characters
(recall that this dictionary is
for students of elementary school and junior high school).
As shown in Fig. \ref{fig:joyo},
it is found that the degree distributions of
the networks of the common-use characters
do not show power-law behavior in the four dictionaries.
These degree distributions
share the features that
there are plateaus in the range of small $k\ (k\lesssim10)$
and decay in large $k\ (k\gtrsim10)$.

\section{Invasion model for selecting the common-use characters}
\begin{figure}[!ht]
\centering
\includegraphics[width=\textwidth]{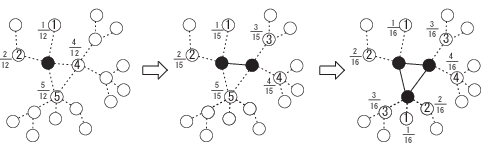}
\caption{K. Yamamoto and Y. Yamazaki}
\label{fig:prob}
\end{figure}

The property of the degree distributions
of the restricted networks shown above is
considered to be caused by a selection process
of the common-use characters.
For this process,
we propose a stochastic model on the `real'
maximal network of each dictionary.
First, we assume that each node in the network has two states;
invaded or uninvaded,
and that all nodes are initially uninvaded.
Then, one node is chosen randomly from the network
and is turned into invaded.
At each time step,
one node $v_i$ is chosen with a probability $p_i$
from all uninvaded nodes $\{v_1,v_2,\cdots,v_n\}$ 
connecting to invaded nodes.
The probability $p_i$ is assumed to be given by
\begin{equation}
p_i=\frac{k_i^\alpha}{\sum_{j=1}^n k_j^\alpha}
\quad (i=1,\cdots,n),
\label{eq:prob}
\end{equation}
where $k_j$ represents a degree of a node $v_j$
and $\alpha$ is a constant, which is determined below.
It is noted that the case $\alpha=0$ corresponds to random growth,
that is, all $v_i$ have equal probability of invasion,
and that the case $\alpha>0$ corresponds to `preferential' growth,
that is, a node of larger degree are invaded more easily
\cite{Barabasi, Krapivsky}.
The invasion process is schematically shown in Fig. \ref{fig:prob}.
In this model,
invaded nodes are regarded as the common-use characters.
The value of $\alpha$ should be positive,
since the common-use characters tend to have large degrees.
Thus, it can be said that this model is
a preferential growth process of an invaded cluster on a network,
and
it is similar to invasion percolation \cite{Wilkinson}
or Eden model \cite{Eden}.

\begin{figure}[!ht]
\centering
\includegraphics[width=\textwidth]{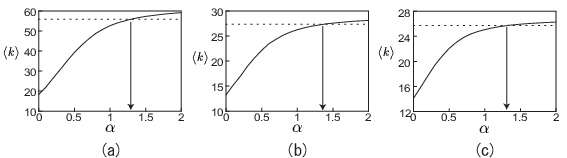}
\caption{K. Yamamoto and Y. Yamazaki}
\label{fig:p-k}
\end{figure}

The process of invasion was performed numerically
until the number of invaded nodes amounted to
the size of the network of the common-use characters
in the dictionaries except {\it Mitsumura}.
Then, we calculated $\ang{k}$ for the
subnetwork of invaded nodes.
To determine $\alpha$,
we require that the average degree $\ang{k}$
of the subnetwork of invaded nodes becomes
almost the same as that of real network of common-use characters.
$\ang{k}$ as a function of $\alpha$
is depicted in Fig. \ref{fig:p-k} in the range $0<\alpha<2$.
From this figure,
it is suggested that $\alpha\approx1.3$ is appropriate
for the three dictionaries.

\begin{figure}[!ht]
\centering
\includegraphics[width=0.8\textwidth]{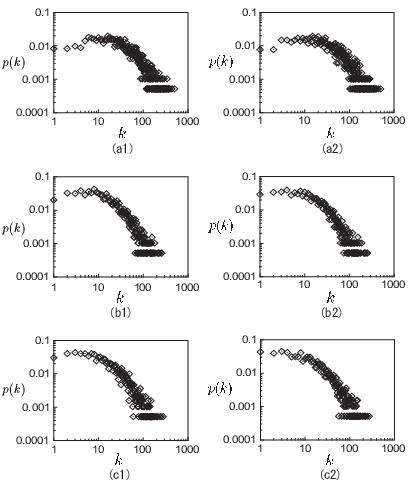}
\caption{K. Yamamoto and Y. Yamazaki}
\label{fig:numerical}
\end{figure}

Numerical results are in good agreement with real networks
as shown in Table \ref{tbl:comparison}.
And Fig. \ref{fig:numerical} shows that
degree distributions obtained from numerical result
are also in good agreement with
those obtained from real networks.

\section{Discussion}
Our analysis has proved that
the network of two-character compounds has both
small-world and scale-free properties.
The possibility of emergence of
the scale-free property seems to be
associated with a fitness model \cite{Caldarelli}.
In the fitness model,
each node $v_i$ in a network has a fitness $x_i$
which is distributed independently and randomly with
a given distribution function $\rho(x)$
(fitness generally represents some kind of
``importance'' or ``sociability'' of nodes).
The edge between $v_i$ and $v_j$ is drawn
with a probability given by $f(x_i,x_j)$ depending on
the fitness of the nodes involved.
And it is known that the fitness model
can produce a power-law degree distribution.

For the network of two-character compounds,
the frequency of use is uneven for each Chinese character:
some characters are used quite frequently
and some characters are used only in particular cases.
And, it is naturally thought that
creation of two-character compounds 
between Chinese characters used more widely
arises more frequently.
Hence, there may be an effect
related to the fitness model
so that the network of two-character compounds
has the scale-free property
(a fitness in this case relates to frequency of use).

We have also found that a network of the common-use characters
is connective, and 
the average degree of the network is larger than
that of the whole network.
The invasion model proposed above
is a simple method to assure connectivity and large degree
of a resultant network.
The model involves one parameter $\alpha$, and
a growth process of invaded cluster depends on the value of $\alpha$.
For positive $\alpha$, nodes of larger degree are assigned
larger invasion probabolity according to Eq. \eqref{eq:prob}.
Hence most nodes of small degree don't join a network
generated from the model, and
power-law behavior in degree distribution vanishes.
Moreover, for appropriate value of $\alpha$, plateau emerges
in a range of small $k$ in the degree distribution.
It is proved that
the value of $\alpha$ is
nearly 1.3 for the three dictionaries in common,
but we have not yet found clear explanation for this universality.

We have confirmed that the network characteristics (Table \ref{tbl:maximal})
and degree distributions (Figs. \ref{fig:degree} and \ref{fig:joyo})
are essentially the same when the edge direction and multiplicity and
self loops are took into account.
We think that
further analysis with direction and multiplicity will provide
more precise structures of the network of two-character compounds.
However, such analysis may be rather linguistic or lexical.
In fact,
direction and multiplicity of edges are
closely related to the individual meanings of characters
and a formation principle of Japanese two-character compounds,
which is classified into nine types
from a grammatical point of view \cite{Nomura}.

\section{Conclusion}
A network constructed by the two-character compounds
in Japanese language has
short path length and high clustering
(Table \ref{tbl:maximal}).
Also the network has power-law degree distribution
(Fig. \ref{fig:degree}),
but a subnetwork restricted to the common-use characters
does not show power-law distribution
(Fig. \ref{fig:joyo}).
Generation of the network of
the common-use characters can be modeled by
an invasion process in which
the invasion probability of nodes with degree $k$
is proportional to $k^\alpha$
(see Eq. \eqref{eq:prob}).
The exponent $\alpha$ is determined by
consistency between real and numerical values of $\ang{k}$
(Fig. \ref{fig:p-k}).
It confirmed that the results obtained from
the model are consistent with real networks quite well
(Table \ref{tbl:comparison}).

%%%%%%%%%%%%%%%%%%%%%%%%%%%%%%%%%%%%%%%%%%%%%%%%%%%%%%%%%%%%%%%%%%%%%
\newpage
\section*{FIGURES \& TABLES}

\begin{figure}[htb]
\centering
\includegraphics[width=\textwidth]{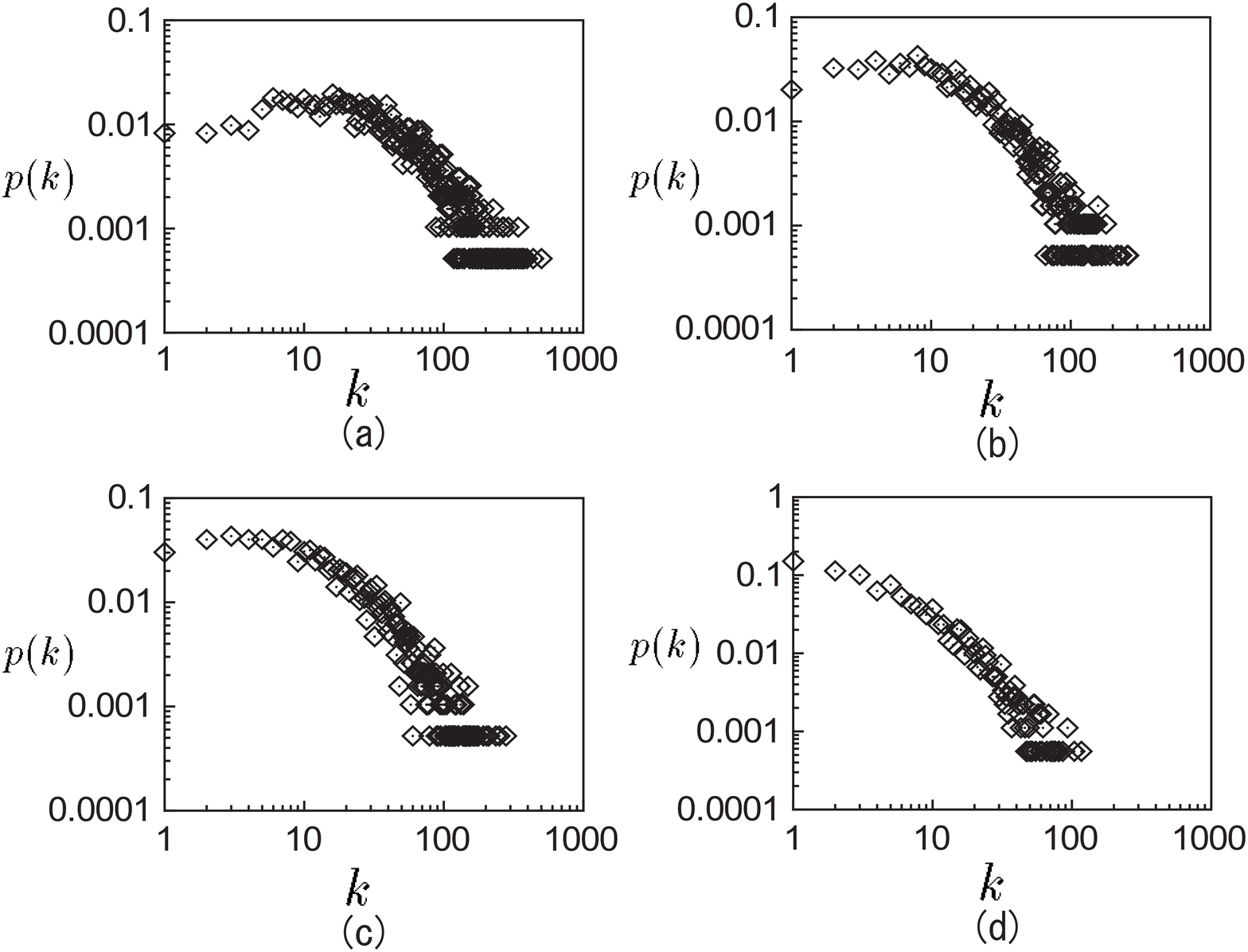}
\caption{K. Yamamoto and Y. Yamazaki}
\label{fig:joyo}
\end{figure}

% figures
\begin{enumerate}
\renewcommand{\labelenumi}{Fig. \arabic{enumi}}
\item % fig1
A part of the network extracted from {\it Kojien}:
(a) original network,
(b) network omitting direction, multiple edges, and self loops

\item % fig2
Degree distribution of the network of each dictionary:
(a) {\it Kojien}, (b) {\it Iwanami},
(c) {\it Sanseido}, and (d) {\it Mitsumura}.
In (a)-(c),
the solid lines show guidelines of power-law behaviors.

\item % fig3
Degree distributions of networks of the common-use characters:
(a) {\it Kojien}, (b) {\it Iwanami},
(c) {\it Sanseido} and (d) {\it Mitsumura}.

\item % fig4
An illustration of an invasion process on a network.
A number on each node indicates a degree of the node,
and a fractional number beside each node indicates 
an invasion probability ($\alpha=1$) at each time step.
The dashed and lines indicate edges of the whole network and
generated subnetwork, respectively.
Black nodes represent that they are invaded.

\item % fig5
Numerical results
to determine the value of $\alpha$.
The solid lines represent the average degree $\ang{k}$
obtained from the model as a function of $\alpha$
(averaging 50 samples),
and dashed lines represent $\ang{k}$
shown in Tabale \ref{tbl:joyo}.
The intersection of solid and dashed lines
indicates $\alpha\approx$ (a) $1.29$ for {\it Kojien},
(b) $1.35$ for {\it Iwanami}, and (c) $1.33$ for {\it Sanseido}.

\item % fig6
Degree distributions of real common-use characters
corresponding to (a1) {\it Kojien}, (b1) {\it Iwanami},
and (c1) {\it Sanseido},
and ones obtained from numerical results
corresponding to (a2) {\it Kojien}, (b2) {\it Iwanami},
and (c2) {\it Sanseido}.
(a1), (b1), and (c1) are identical to Fig. \ref{fig:joyo}
(a)-(c).

\end{enumerate}

\bigskip

% tables
\begin{enumerate}
\renewcommand{\labelenumi}{Table \arabic{enumi}}

\item % table1
The characteristics of the maximal cluster
in a network of two-character compounds.
$\ang{k}$, $\ell$, $D$, and $C$ denote
average degree, mean path length, diameter, and
clustering coefficient, respectively.
$C_{rand}$ represents the averaged clustering coefficient of the
 $ 50$ random networks of the same size in nodes and edges.

\item % table2
The characteristics of the network of common-use characters.

\item % table3
Comparison between real networks of common-use characters
and numerical results ($\alpha=1.3$).
Numerical results are obtained by averaging $50$ samples.

\end{enumerate}

%%%%%%%%%%%%%%%%%%%%%%%%%%%%%%%%%%%%%%%%%%%%%%%%%%%%%%%%%%%%%%%%%%%%%
\newpage

\newpage

\clearpage

\newpage

\newpage

\newpage

%%%%%%%%%%%%%%%%%%%%%%%%%%%%%%%%%%%%%%%%%%%%%%%%%%%%%%%%%%%%%%%%%%%%%%
\newpage
\begin{table}[h]
\caption{K. Yamamoto and Y. Yamazaki}
\label{tbl:maximal}
\centering
\begin{tabular}{ccccccccc}
\hline\hline
Dictionary & Nodes & Edges & $\ang{k}$ & $\ell$ & $D$ & $C$ & $C_{rand}$ & $\gamma$\\
\hline
{\it Kojien}    & 5458 & 74617 & 27.3 & 3.14 & 10 & 0.138 & 0.00501 & 1.04 \\
{\it Iwanami}   & 3904 & 32150 & 16.5 & 3.31 & 10 &0.085 & 0.00424 & 1.04 \\
{\it Sanseido}  & 3444 & 28358 & 16.5 & 3.32 & 9 &0.086 & 0.00483 & 1.05 \\
{\it Mitsumura} & 1799 & 9054  & 10.1 & 3.42 & 8 &0.059 & 0.00255 & --- \\
\hline\hline
\end{tabular}
\end{table}

\newpage
\begin{table}[htb]
\caption{K. Yamamoto and Y. Yamazaki}
\label{tbl:joyo}
\centering
\begin{tabular}{ccccccc}
\hline\hline
dictionary & Nodes & Edges & $\ang{k}$ & $\ell$ & $D$ & $C$\\
\hline
{\it Kojien} & 1940 & 54181 & 55.9 & 2.32 & 5 & 0.172 \\
{\it Iwanami} & 1933 & 26419 & 27.3 & 2.67 & 6 & 0.111 \\
{\it Sanseido} & 1921 & 24726 & 25.7 & 2.73 & 7 & 0.114 \\
{\it Mitsumura} & 1799 & 9054 & 10.1 & 3.42 & 8 & 0.059 \\
\hline\hline
\end{tabular}
\end{table}

\newpage
\begin{table}[!h]
\caption{K. Yamamoto and Y. Yamazaki}
\label{tbl:comparison}
\centering
\begin{tabular}{c@{\hspace{3\tabcolsep}}ccc@{\hspace{3\tabcolsep}}ccc}
\hline\hline
\multirow{2}{*}{dictionary} & \multicolumn{3}{c}{Real networks} &
\multicolumn{3}{c}{Numerical results}\\
 & $\ang{k}$ & $\ell$ & C & $\ang{k}$ & $\ell$ & C\\
\hline
{\it Kojien} & 55.9 & 2.32 & 0.172
	& $56.0\pm0.8$ & $2.32\pm0.01$ & $0.175\pm0.004$\\
{\it Iwanami} & 27.3 & 2.67 & 0.111
	& $27.2\pm0.2$ & $2.68\pm0.01$ & $0.109\pm0.005$\\
{\it Sanseido} & 25.7 & 2.73 & 0.114
	& $25.7\pm0.2$ & $2.73\pm0.02$ & $0.109\pm0.004$\\
\hline\hline
\end{tabular}
\label{tbl:numerical}
\end{table}

\end{document}